\documentclass[lettersize,journal]{IEEEtran}
\usepackage{amsmath,amsfonts}
\usepackage{array}
\usepackage{textcomp}
\usepackage{stfloats}
\usepackage{url}
\usepackage{verbatim}
\usepackage{graphicx}
\usepackage{algorithm}
\usepackage{algpseudocode}
\usepackage{subfigure}
\usepackage{caption}
\usepackage{subcaption}
\usepackage{float}
\usepackage{epsfig}
\usepackage{xcolor}
\usepackage{tikz}
\usetikzlibrary{trees, shadows}
\usepackage{tikz-qtree}
\usepackage{tablefootnote}
\usepackage [para,online,flushleft]{threeparttable}
\usepackage{hhline}
\usepackage{colortbl}
\usepackage{amssymb}
\usepackage{comment}
\usepackage{pdflscape}
\usepackage{svg}
\usepackage{pdfpages}
\usepackage{soul}
\usepackage{mdframed}
\usepackage{adjustbox}
\usepackage{flushend }
\usepackage{orcidlink}
\usetikzlibrary{trees,positioning,shapes,shadows,arrows}

\definecolor{fashionfuchsia}{rgb}{0.96, 0.0, 0.63}

\def\BibTeX{{\rm B\kern-.05em{\sc i\kern-.025em b}\kern-.08em
    T\kern-.1667em\lower.7ex\hbox{E}\kern-.125emX}}
\usepackage{balance}
\begin{document}
\title{GLow - A Novel, Flower-Based Simulated Gossip Learning Strategy}
\author{Aitor Belenguer\IEEEauthorrefmark{1}~\orcidlink{0000-0003-2079-0137}, Jose A. Pascual\IEEEauthorrefmark{1}~\orcidlink{0000-0001-5355-6537}, Javier Navaridas\IEEEauthorrefmark{1}~\orcidlink{0000-0001-7272-6597}
\\\IEEEauthorblockA{\IEEEauthorrefmark{1}University of the Basque Country UPV/EHU
\\\{aitor.belenguer, joseantonio.pascual, javier.navaridas\}@ehu.eus}
}


\maketitle

\begin{abstract}
Fully decentralized learning algorithms are still in an early stage of development. Creating modular Gossip Learning strategies is not trivial due to convergence challenges and Byzantine faults intrinsic in systems of decentralized nature. Our contribution provides a novel means to simulate custom Gossip Learning systems by leveraging the state-of-the-art Flower Framework. Specifically, we introduce GLow, which will allow researchers to train and assess scalability and convergence of devices, across custom network topologies, before making a physical deployment. The Flower Framework is selected for being a simulation featured library with a very active community on Federated Learning research. However, Flower exclusively includes vanilla Federated Learning strategies and, thus, is not originally designed to perform simulations without a centralized authority. GLow is presented to fill this gap and make simulation of Gossip Learning systems possible. Results achieved by GLow in the MNIST and CIFAR10 datasets, show accuracies over 0.98 and 0.75 respectively. More importantly, GLow performs similarly in terms of accuracy and convergence to its analogous Centralized and Federated approaches in all designed experiments.
\end{abstract}


\section{Introduction} \label{intro}

\IEEEPARstart{T}{he} number of IoT devices has surged in recent decades, driven by smart cities, wearables, self-driving cars and automation. The data generated by these devices is crucial for training complex ML/DL models, traditionally done in centralized datacenters (CNL). However, due to scalability, privacy and resource concerns, new paradigms like Federated Learning (FL)~\cite{BrendanMcMahan2017} have emerged, offering reduced communication costs and privacy benefits. Gossip Learning (GL)~\cite{decentr_gl} further addresses central authority issues by enabling fully decentralized communication among neighbors.

GLow is a novel simulated GL strategy, built on top of the Flower Framework~\cite{flower}. That allows researchers to test custom GL systems in terms of convergence, scalability and performance. Flower is a state-of-the-art framework developed to support FL studies by implementing strategies based on a centralized aggregation server. Nevertheless, we envision a scenario where different nature IoT devices are connected to each other following a decentralized interconnection scheme -- without an aggregation server. GLow leverages Flower capabilities to enable fully decentralized Gossip Learning, in which each agent holds a dedicated model and benefits from neighbor parameter interchange. An important feature of GLow is that network topology and agents behavior can be customized, providing an enormous degree of freedom in the design of decentralized systems. In particular, we propose a novel methodology where agents can be instructed to behave in a predefined fashion, which can be leveraged to ensure the correct evolution of the decentralized learning system.

As a case of study, we carry out experiments using GLow with a series of simple topologies and compare them against analogous CNL and FL approaches, discussing the trade-offs involved with each technology. Our experimental set-up includes MNIST and CIFAR10 datasets, variable number of local epochs and various configurations with different number of agents and connectivity degrees: 8 agents across 5 topologies, from disconnected to fully connected networks, and 16 agents across 9 topologies, ranging from no connection to full connectivity. Results show up to 0.98 accuracies in MNIST and up to 0.75 in CIFAR10 datasets. As described in the following sections, GLow is a well-suited, modular and decentralized simulated GL strategy, which is shown to be competitive when compared to CNL and FL approaches.
\section{State of the art} \label{state_art}
\subsection{Self Learning and Centralized Learning}
In Self Learning (SL), each agent trains a model exclusively relying on local data. It can be a good and simple learning approach, when the local database is large enough and the device is placed in a remote location with no access to the network or under restrictive policies that force it to work in offline mode. However, receiving data or other model parameters from the network can potentially benefit the creation of stronger models. Specifically, learning from other agents is of special interest in scenarios where the nature of data changes constantly or stream learning is performed, e.g., Anomaly Detection, Network Intrusion Detection Systems~\cite{borja} and so on.   
In order to overcome the SL scenario, working with multiple IoT agents with a similar task allows sharing dataset instances in a raw way to a centralized authority, that is when traditional Datacenter or Centralized Learning (CNL) appears. However, new privacy, reliability and scalability challenges emerge: (1) sharing highly sensitive data through the network can violate privacy constraints, (2) relying on a particular centralized authority to perform all the training, makes the system vulnerable against a weak spot, (3) sharing data from a cross-device~\cite{surv2} environment with a large number of devices to a datacenter can cause network throughput issues, even to the point of saturating the whole network.

\subsection{Federated Learning}
Federated Learning (FL) enables multiple parties to jointly train an ML model without exchanging local data. FL was first proposed by B. McMahan et al. (FedAVG)~\cite{BrendanMcMahan2017}, involving distributed systems, ML and privacy research areas. Currently, FL has become a hot-topic of research and different Federated Learning Systems have emerged. Depending on their nature and objectives can be classified following the scheme proposed by Q. Li et al~\cite{surv2}: (1) data partitioning method, (2) trained ML model, (3) privacy mechanism, (4) communication architecture, (5) scale of federation, (6) motivation of federation.
Although FL introduces important features in terms of privacy and scalability compared to traditional datacenter learning, it involves the use of a centralized authority to perform parameter averaging. During the FL training stage, each agent trains its local model with local dataset instances and sends only learned parameters to an orchestration authority (server) -- privacy mechanisms can be included at this point, e.g., Laplacian Noise addition~\cite{S48}. The server is in charge of applying an aggregation algorithm and sending those aggregated parameters back to the agents. Beyond FL, custom aggregation algorithms and Attention Mechanisms can be applied~\cite{Liu2021}. Nevertheless, the usage of a server to orchestrate the learning is translated into a potential weak-point susceptible to be targeted by attackers as well as could limit scalability in scenarios where a large number of agents, sharing their local parameters, are involved.

\subsection{Gossip Learning}
Gossip Learning (GL) is a decentralized learning method in which different agents interchange local parameters with their neighbors (peers) and perform the aggregation of the models asynchronously -- without needing an orchestration server. Although Ormandi et al.~\cite{ormandi} introduced the idea of GL for classification using linear models back in 2013, research about fully decentralized algorithms has not attracted as many researches as vanilla FL, mostly because of the additional convergence and orchestration challenges GL entails. However, L. Yuan et al.~\cite{Yuan2023} introduce a different point of view of FL, dividing it into Centralized FL (CFL) and Decentralized FL (DFL), based on the existence of an aggregator server. Moreover, a taxonomy of DFL systems is proposed in which GL is considered a communication protocol and other aspects such as iteration order and network topology are taken into account. Additionally, E. Martinez et al.~\cite{mbeltran} claim the lack of contributions gathering information about existing DFL frameworks and classify them based on how mature and customizable those frameworks are. It is remarkable that, in spite of mentioning frameworks such as TensorFlow Federated (TFF), PySyft or FedML, there is no research involving DFL using Flower~\cite{flower}, even when it is a consolidated framework in the FL community.

\subsection{Flower Framework}
Flower is a comprehensive \textit{open-sourced} FL framework that offers new facilities to carry out large-scale FL experiments and considers richly heterogeneous FL device scenarios~\cite{flower}. It unlocks scalable algorithmic research as well as system level aspects. In other words, it allows the transition from experimental research in simulation to system research in heterogeneous edge devices. D. Beutel et al.~\cite{flower} present a comparison among different FL frameworks, stating that Flower provides a larger feature set in comparison to TFF, Syft, FedScale and LEAF. Although Flower is interesting in terms of simulation, scalability and multi-platform deployment properties; current research and designed algorithms are CFL oriented (e.g., FedAVG, FedProx and so on). To the best of our knowledge, a research line involving DFL in Flower is not fully covered by the state-of-the-art.

\subsection{Related Work}
Although FL is a hot-topic of research, contributions to Decentralized FL strategies are limited.  Roy et al.~\cite{braintorrent} present BrainTorrent, a dynamic peer-to-peer environment, and carry out a proof-of-concept study that performs medical image segmentation. The authors propose a fully connected network, where each agent maintains a vector containing its own and the last versions of the models used during the merging step. Another interesting contribution is Fedstellar~\cite{fedstellar}, which proposes a novel platform to train FL models in a centralized, decentralized and semi-decentralized fashion, across physical or virtualized devices. The platform is custom-made, merging PyTorch, asynchronous sockets and Docker containers. Fedstellar presents a robust comparison against other literature solutions, including BrainTorrent, using MNIST and CIFAR10 datasets to assess system performance.

L. Chen et al.~\cite{Chen2024} propose the usage of quantization methods to improve DFL convergence. Specifically, the authors suggest using Lloyd-Max algorithm to minimize quantization distortion by adjusting quantization levels -- again, system performance is assessed using MNIST and CIFAR10 datasets. Another alternative is GossipFL~\cite{gossipfl} which presents a novel sparsification algorithm, built on FedML framework, to enable each agent to communicate with just one peer while better using bandwidth resources.

The number of approaches that try to adapt the Flower framework to support DFL scenarios is scarce. The only one we are aware of is introduced by Y. Kanamori et al.~\cite{Kanamori2023} which developed an asynchronous FL framework for decentralized systems. The authors take advantage of gRPC (Google Remote Procedure Calls) present in Flower communication layer to operate at agent message passing level. This approach differs from ours in that it focuses on system deployment rather than on simulated environments.
Their ultimate objective is to understand how the order of learning and aggregation affects the performance of DFL and their evaluation tests various topology-agnostic decentralized algorithms in their framework. In contrast, our approach is more interested in assessing the effects of agent topology, density of communications, etc. Furthermore, our simulation-oriented strategy enables decentralized learning research while preserving Flower framework core unaltered. Specifically, we focus on simulating fully decentralized systems by proposing a custom Flower strategy from an upper level -- without interfering with internal communication layers.


\section{Designed Strategy} \label{design_str}
Flower does not support a fully decentralized aggregation mechanism able to orchestrate multiple network agents and make them converge without a centralized authority. Specifically, Flower simulation engine is not originally designed to deploy agents as servers and clients at the same time -- following a fully decentralized scheme.
In this work, we propose GLow which aims at bridging this gap by dispensing with the need of having a centralized server.
The idea behind GLow is to perform a simulated learning algorithm that allows each agent to learn from its neighbors. 

GLow is implemented so that during each iteration a given agent is designated as the server (head). Selected head will train its local model using local dataset instances and aggregate it with updated model parameters received from its neighbors. A trade-off between convergence speed and scalability is taken into account to assess the performance of the system -- usually constrained by the connectivity graph among agents.

The decentralized aggregation algorithm is designed following Flower simulation guidelines~\cite{flower}, maintaining modularity by modifying only upper-level classes \textit{client, server, model, dataset} and creating a \textit{custom strategy} \footnote{A slight modification in the internal \textit{aggregate} class of server has been made to allow the existence of agents with no local data -- avoiding the execution to crash.}. In particular, the vanilla server parameter spread from the server to the agents as in FedAVG is not performed. The main reason is because in a fully decentralized asynchronous scenario, agents will constantly be asking their neighbors for their updated model parameters as well as performing local training -- each agent will be in charge of asking neighbors about updated information. Therefore, a server spreading parameters is not needed, neither a multicast from the head to the neighbors is performed -- reducing the number of messages sent through the network.

\begin{algorithm}
\caption{GLow Simulation Procedure}\label{alg:cap}
\begin{algorithmic}[1]
\Require Head model weights $W_H^k$ for selected head agent $k$, Local model weights $W_L^i$ for each neighbor agent $i$ of head; $total\_rounds, E \in \mathbb{Z}^+$
\Ensure Aggregated head model weights $W_H^{k'}$
\For{$each\ iteration\ in\ 1\ to\ total\_iterations\ $}
    \State $k = iteration \mod{K}$
    \State $ W^{k'}_{L} = $ agent $k$ performs local training $E$ times
    \State $k$ gets current $W_L^{i'}$ of each Neighbour agent $\cal{N}$$_k$ 
    \State $W_H^{k'} = \sum_{i \in \cal{N}}W_L^{i'}$
    
\EndFor
\end{algorithmic}
\end{algorithm}

Algorithm~\ref{alg:cap} shows how GLow simulation procedure is performed: (1) the agent head is selected in a round-robin fashion by using the modulus of current iteration~\footnote{Iterations are a consequence of simulating decentralized system communication rounds (i.e., represent the sum of all asynchronous communication rounds of every agent).} -- other strategies such as random or priorities would be straightforward to implement, (2) the selected head performs local training using local instances for $E$ epochs, (3) the Head agent asks his neighbors $\cal{N}$$_k$ for their current weights $W_L^{i'}$, (4) aggregation (weighted average) among locally trained and neighbor parameters is performed, updating head model $W_H^{k'}$.



In order to select how agents are connected, GLow includes a Topology Generator (TG) tool which allows assessing system behavior in different decentralized scenarios. The output generated by TG is completely compatible with GLow input requirements. Moreover, it makes the process of evaluating system convergence easier by automatically generating topologies with different connectivity degrees. In the experiments below, TG creates topologies from disconnected to fully connected by connecting each agent to the closest $n$ neighbors in a ring (undirected graph) -- See Figure~\ref{fig:topologies}. 
Note, however, that GLow offers a set of methods for generating other special topologies such as: \textit{chain, star chain, ring, ring chain} and \textit{fully connected}~\footnote{Figures: https://github.com/AitorB16/GLow/tree/master/img}.
Additionally, as will be explained more thoroughly in Section~\ref{Sp_Agents}, our methodology allows for special agents that enhance experimental flexibility. For this reason, all methods above accept a list of disconnected agents that will always perform SL and a list of agents with no local data that will only improve their model by aggregating it with the parameters received from their neighbors.
\section{Experimental Setup} \label{E_Setup}
GLow is a novel Flower-based GL strategy that needs to be compared against other learning paradigms to assess its feasibility and benefits. Although creating a fair testing scenario among systems of different nature is challenging, a common frame is designed to analyze the trade-off of using CNL, FedAVG (FL) and GLow (GL) approaches.

\begin{figure}
\begin{tabular}{cc}
  \includegraphics[width=40mm]{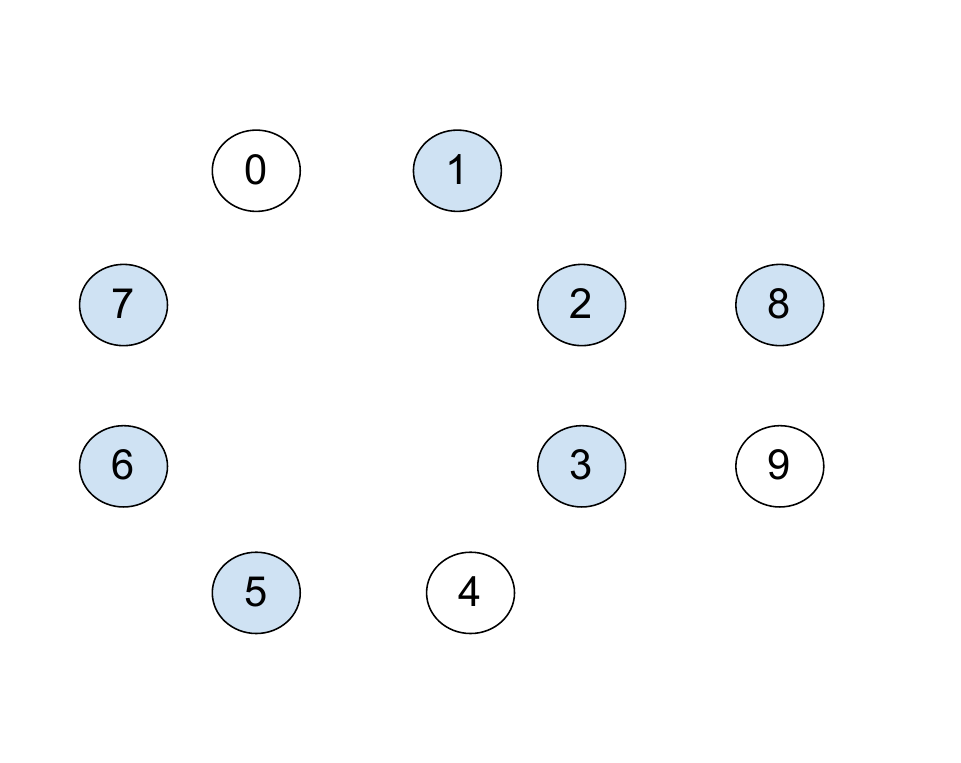} &   \includegraphics[width=40mm]{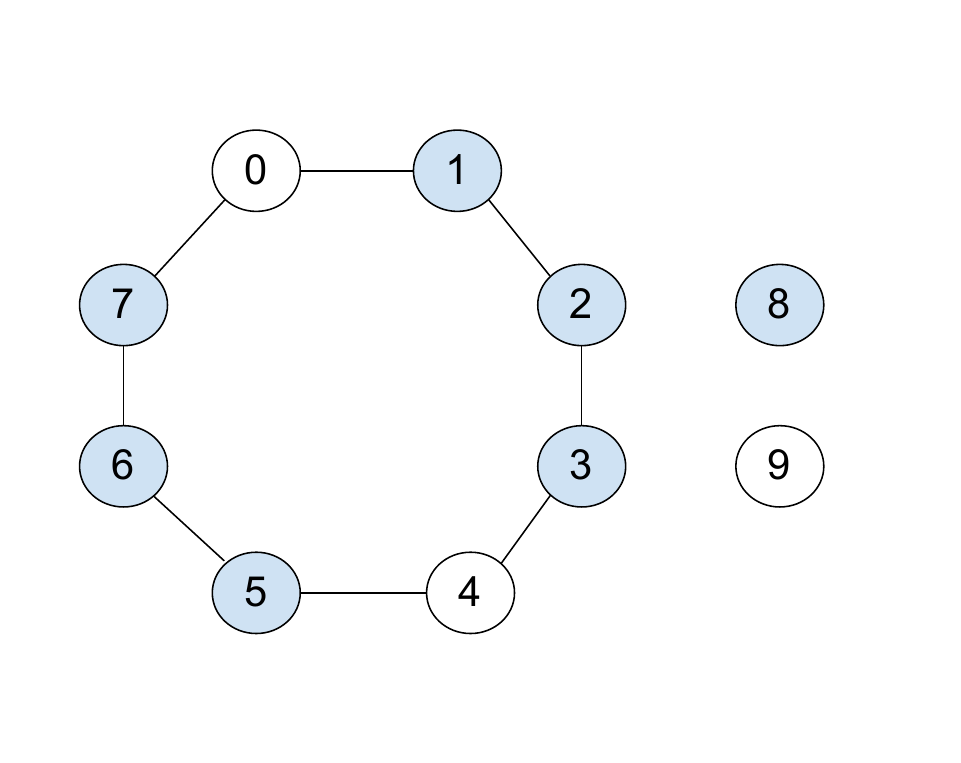} \\
(a) Topo 0 - Disconnected & (b) Topo 2 - Ring\\[5pt]
 \includegraphics[width=40mm]{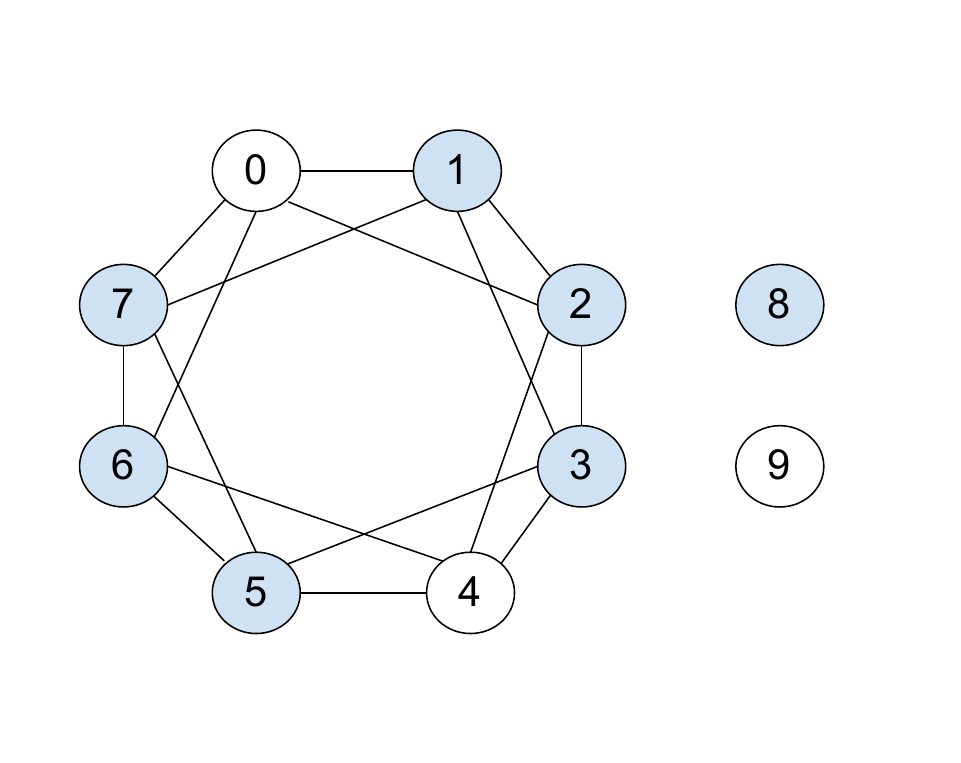} &   \includegraphics[width=40mm]{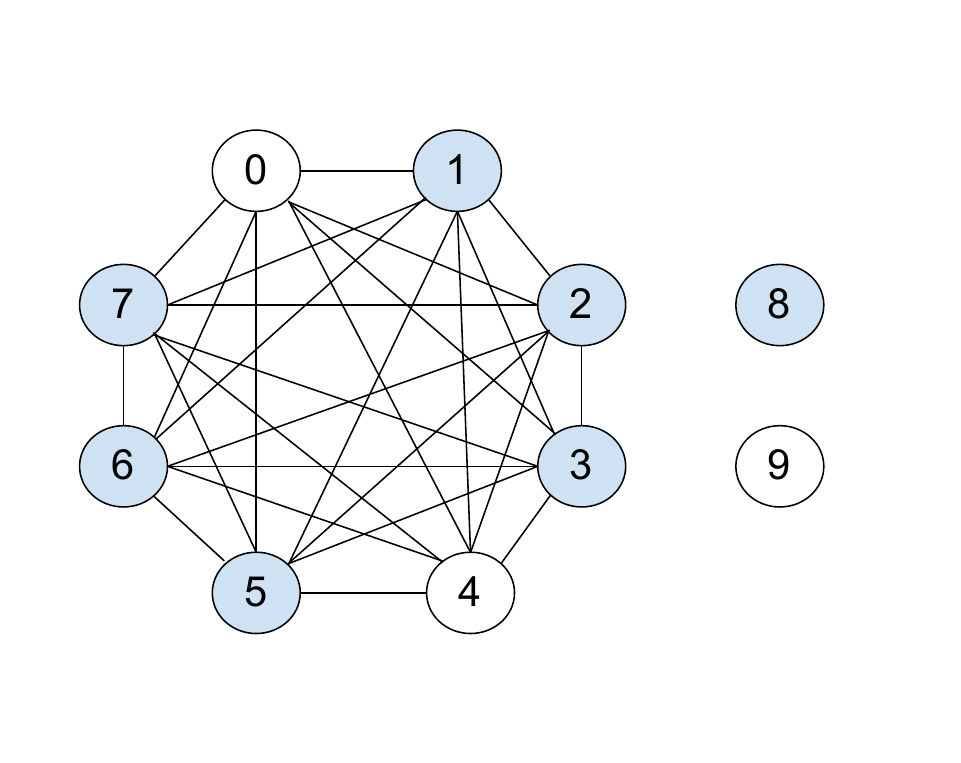} \\
(c) Topo 4 - Double ring & (d) Topo 7 - F. connected\\[5pt]
\end{tabular}
\caption{Topologies in 8+2 agent configuration, from a fully disconnected graph to a fully connected one -- connecting each agent to the next two neighbors respectively. Agents with local data are represented with blue bullets and without local data with white bullets.}
\label{fig:topologies}
\end{figure}

In order to assess GLow performance, two different setups are prepared under an image classification task. The convergence of the system is tested with 8 and 16 agents. Moreover, experimentation is performed with MNIST~\cite{MNIST} and CIFAR10~\cite{cifar} datasets to have a solid test bench. We analyze how the interconnection degree impacts in each agents' learning performance by launching experiments from a disconnected (Self Learning) to a fully connected graph scenario. In the proposed scenarios, illustrated in Figure~\ref{fig:topologies}, each topology progressively adds new links to agents by connecting them to the next two closest agents. In other words, 2 new undirected edges are added per agent in each topology; each agent will be connected to 0 neighbors in topology 0, to 2 neighbors in topology 2, ..., to \textit{n} neighbors in topology \textit{n}. Hence, in the 8-agent case, 5 topologies are studied from a completely disconnected to a fully connected graph. Similarly, in the 16-agent case, 9 topologies are studied from a completely disconnected to a fully connected graph. 

In the creation of the mentioned common benchmark for assessing GLow, two additional versions are created based on: (1) CNL learning, where the training is carried out over the whole dataset in a single server and (2) FL, using vanilla FedAVG~\cite{BrendanMcMahan2017} algorithm (splitting the dataset in an IID way among all the agents). Although the performance of CNL and FL versions is expected to be higher (due to their simplicity against fully decentralized), the benefits of using GLow makes it potentially more attractive in cross-device scenarios -- as mentioned in Section~\ref{state_art}. 

\subsection{Datasets and models}
MNIST and CIFAR10 datasets are selected due to their wide usage in state-of-the-art for testing convergence in distributed learning systems. Moreover, using these image classification datasets gives a solid background to our strategy due to the number of available models and acceptance by the community. Thus, evaluated strategies for designed systems: CNL, FL and GLow will be executed twice, each time on both datasets. As described in the following lines, the architecture of the models used in the image classification task is maintained in the three systems. 

MNIST~\cite{MNIST} dataset contains 60000 training and 10000 testing 28$\times$28px grayscale images of handwritten digits (characters `0' to `9'). Therefore, a random guesser should achieve a mean accuracy of around 0.1. LeNet is selected as the main model to be trained following the architecture proposed by Lecun et Al.~\cite{le_net}. 


CIFAR10~\cite{cifar} dataset contains 60000 training and 10000 testing 32$\times$32px images of 10 different classes (airplane, automobile, bird, cat, deer, dog, frog, horse, ship, truck). A random guesser should achieve a mean accuracy around 0.1. As above, we selected a variant of the LeNet architecture working with 32x32 images as proposed by Lecun et Al.~\cite{le_net}. 


\subsection{Special Agents} \label{Sp_Agents}
In this work, we introduce a novel methodology for assessing the correct behavior of a fully decentralized learning system: special agents which behave in a predictable way and serve as control agents that will remain stable during simulation. Therefore, if special agents start running unpredictably, simulation will need to be re-assessed. Table~\ref{table_agents} clusters each agent by \textit{Code} and \textit{Color Group} to make results more representative in Figures~\ref{fig:MNIST_8_32_3} and \ref{fig:CIFAR_8_32_3}. In particular, we consider the following \textit{expected behaviors}: (1) agents with no local data (E) and disconnected from the network (D) are expected to behave as random guessers (ED), (2) agents with local data and disconnected (D -- performing SL) are expected to behave similar or worse than connected agents with local data, (3) agents with no local data but connected to their neighbors (E) are expected to learn from other agents and achieve certain degree of convergence. The remaining agents have data and are connected (R -- Regular).

\begin{table*}[!ht]
\centering
\caption{Summary of existing GLow agents.}
\begin{threeparttable}[t]
\begin{tabular}{l||c|c|c||} 
Setup & Code\tnote{a} & Color Group & Agent IDs \\ 
\hline \hline
8+2       &    R &    Blue-Purple & 1, 2, 3, 4, 5, 6, 7\\
8+2       &    D &    Light Green & 8\\
8+2       &    E &    Orange-Yellow & 0, 4\\
8+2       &    ED &   Red-Dark Red & 9\\
\hline \hline
16+4       &   R &    Blue-Purple & 1, 2, 3, 4, 6, 7, 8, 9, 11, 12, 13, 14, 15\\
16+4       &   D &    Light Green- Green & 16, 17 \\
16+4       &   E &    Orange-Yellow & 0, 5, 10\\
16+4       &   ED &   Red-Dark Red & 18, 19\\
\hline
\end{tabular}
\begin{tablenotes}
    \item[a] R: Regular (connected and with local data), D: Disconnected (SL), E: Empty (no local data), ED: E+D (random guess) -- connectivity applies from topology 2 on.
   \end{tablenotes}
    \end{threeparttable}
\label{table_agents}
\end{table*}

As shown in Figure~\ref{fig:topologies}, in designed experiments, agents with no data are displayed with white bullets. A set of random agents are selected as agents with no local instances to test how information is learned from the network -- through neighbor agents. Moreover, the concept of disconnected agents is introduced in all the executions -- creating a baseline of SL agents against interconnected ones. Specifically, in the 8 agent scenario, the ones with no local data are (0, 4, 9) and the disconnected ones are (8, 9); making a total of 10 agents -- 8 connected + 2 disconnected. On the other hand, in the 16 agent scenario, the ones with no local data are (0, 5, 10, 18, 19) and the disconnected ones are (16, 17, 18, 19); making a total of 20 agents -- 16 connected + 4 disconnected. It has to be clarified that for the FL experiments, agents with no local data are maintained, whereas there are no disconnected ones. As a consequence, FL \textit{Agent Number} notation in Table~\ref{table_res} is provided within a single number -- GLow is displayed as a sum of connected + disconnected.



\subsection{Configurations}
To ascertain the effects of agent scale in GLow, two experimentation scenarios are carried out with 8+2 and 16+4 agents respectively.
In the experiments we refer to communication rounds as the total number of times each agent is designated as head and performs parameter averaging. Therefore, the total number of iterations used in Algorithm~\ref{alg:cap} is the number of agents times the number of communication rounds.
Depending on the selected dataset and its complexity, the number of communication rounds is set differently to ensure proper convergence: 24 communication rounds per agent as head in MNIST, whereas 101 communication rounds per agent as head in CIFAR10. Section~\ref{design_str} shows how an agent is selected as head (operating as server and client) while neighbor parameters are aggregated with the local ones in each communication round. 

Both experimentation scenarios are launched 5 times each with a different number of local training epochs per agent; 2, 4, 8, 16 and 32. Thus, every time an agent is selected as head, it performs a local training with a given number of epochs and then requests parameters to its neighbors for being aggregated into its own model.
\section{Results} \label{results}

We now move on to the evaluation of our case study, where we assess the feasibility of the GLow strategy by comparing it with CNL and FL. 
We carried out an extensive evaluation\footnote{The full results from our experiments are publicly available at: \url{https://github.com/AitorB16/GLow}} with many topologies and numbers of epochs, see Section~\ref{E_Setup}. Nonetheless, for the sake of brevity and clarity, we will focus on a few representative results that highlight our main findings. 

First, we show the results obtained by CNL and vanilla FedAVG systems as our baseline. Then, we use GLow to analyze how the degree of connectivity of each topology affects the overall performance. For simplicity, we focus on accuracy and loss evolution as our main metrics. Table~\ref{table_res} summarizes results obtained in each of the three systems by showing the average accuracies obtained after running each experiment. In the case of GLow, as it is a fully decentralized strategy that includes disconnected nodes, we only take into account the local accuracies of the connected agents (E and R) -- D and ED agents are not part of the GL group and therefore their results are discarded. In Table~\ref{table_res}, results for GLow are those for Topology 4 (double ring). The reason behind is that this is the first topology in terms of degree of connectivity achieving similar levels of convergence in all E and R agents. Note that this means GLow is handicapped in two ways. First, GLow's model is trained with a subset of the data used by its homologues. This is because part of the dataset resides in the disconnected agents. Secondly, GLow's system incorporates agents without data which are detrimental to the accuracy of the model.

\begin{table*}[!ht]
\centering
\caption{Summary of obtained results in CNL, FL and GLow systems.}
\begin{threeparttable}[t]
\begin{tabular}{l||c|c|c|c||c||} 
System & Dataset & Agent Number & Communication Rounds & Local Epochs\tnote{a} & Average Accuracy\tnote{b} \\ 
\hline \hline
CNL       & MNIST              &  - &    - & 24 & 0.989 \\
CNL       & CIFAR10              &  - &    - & 101 & 0.789  \\
\hline \hline
FL       & MNIST              &  10 &    24 & 32 & 0.985 \\
FL       & CIFAR10              &  10 &    101 & 32 &  0.791 \\
FL       & MNIST              &  20 &    24 & 32  & 0.986 \\
FL       & CIFAR10              &  20 &    101 & 32 & 0.778 \\
\hline \hline
GLow       & MNIST              &  8+2 &    24 & 32 & 0.987 \\
GLow       & CIFAR10              &  8+2 &    101 & 32 & 0.754 \\
GLow       & MNIST              &  16+4 &    24 & 32 &  0.971 \\
GLow       & CIFAR10              &  16+4 &    101 & 32 & 0.683 \\
\hline
\end{tabular}
\begin{tablenotes}
    \item[a] There are no \textit{Communication Rounds} in CNL; \textit{Local Epochs} will be the total epoch number of the system.
    \item[b] Results of disconnected agents are not taken into account in GLow avg accuracy computation.
   \end{tablenotes}
    \end{threeparttable}
\label{table_res}
\end{table*}

\subsection{CNL and FL}

\begin{figure*}[!ht]
    \centering
    \begin{subfigure}
        \centering
        \includegraphics[width = .45\textwidth]{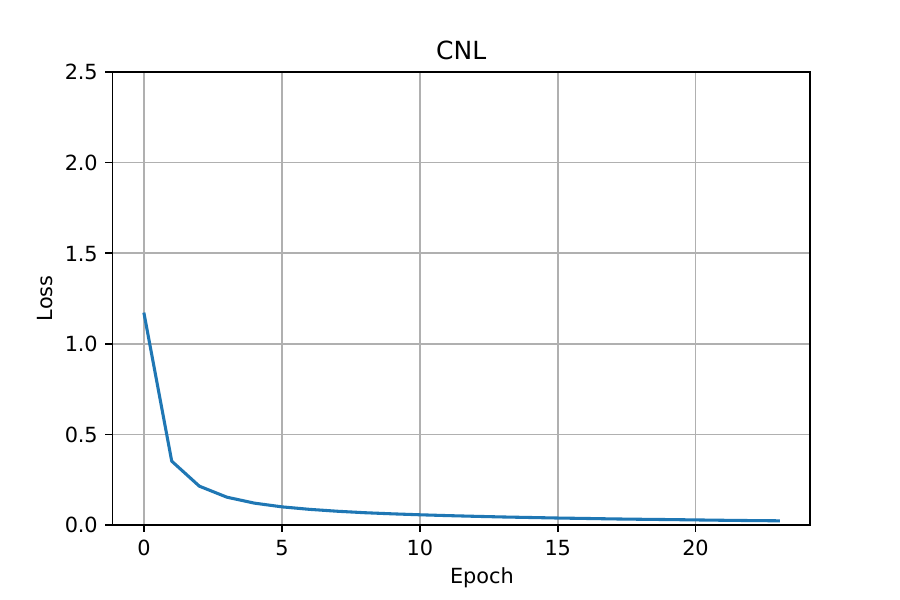}
   \end{subfigure}
    \begin{subfigure}
        \centering
        \includegraphics[width = .45\textwidth]{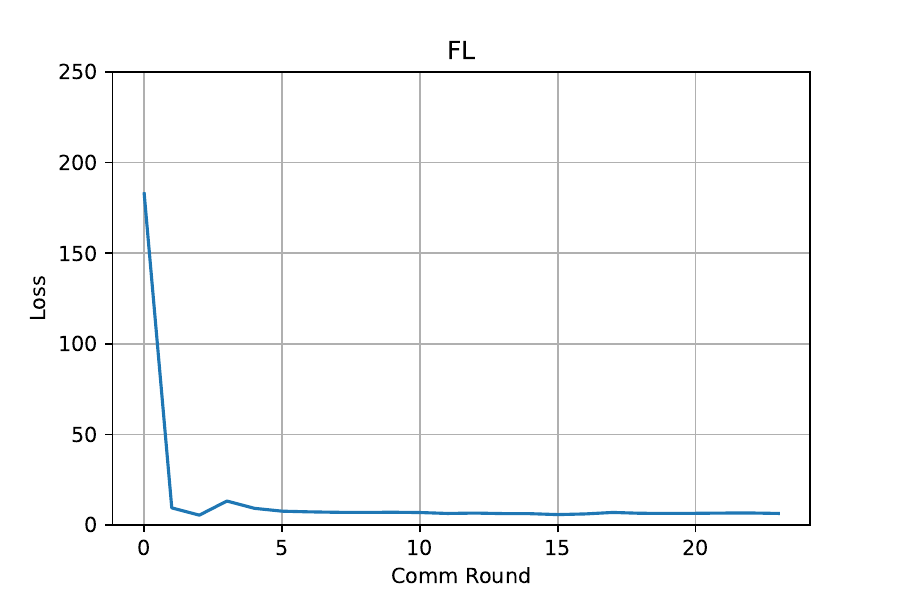}
    \end{subfigure}
    \caption{Loss evolution of CNL (left) and FL (right) systems in the MNIST dataset.}
    \label{fig:FL_MNIST}
\end{figure*}

\begin{figure*}[!ht]
    \centering
    \begin{subfigure}
        \centering
        \includegraphics[width = .45\textwidth]{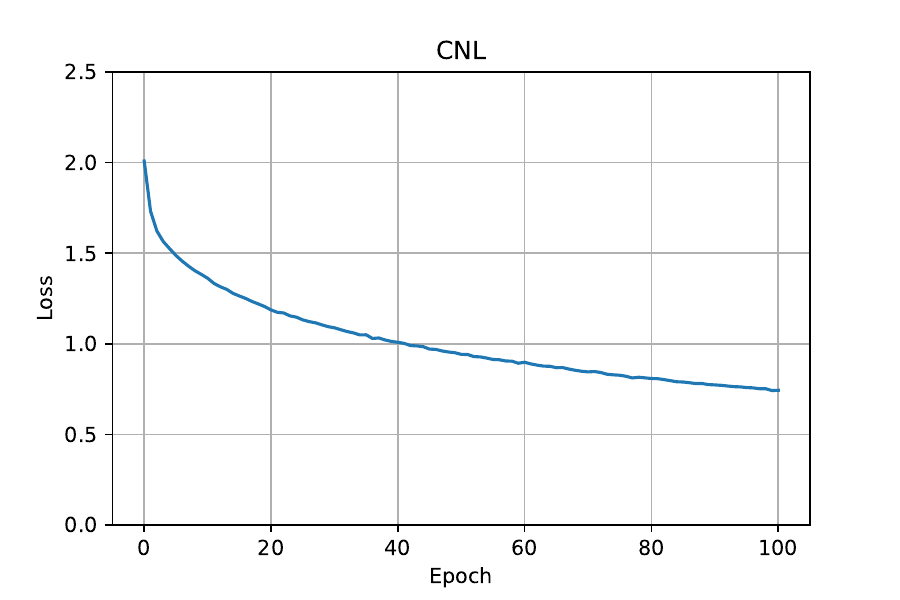}
   \end{subfigure}
    \begin{subfigure}
        \centering
        \includegraphics[width = .45\textwidth]{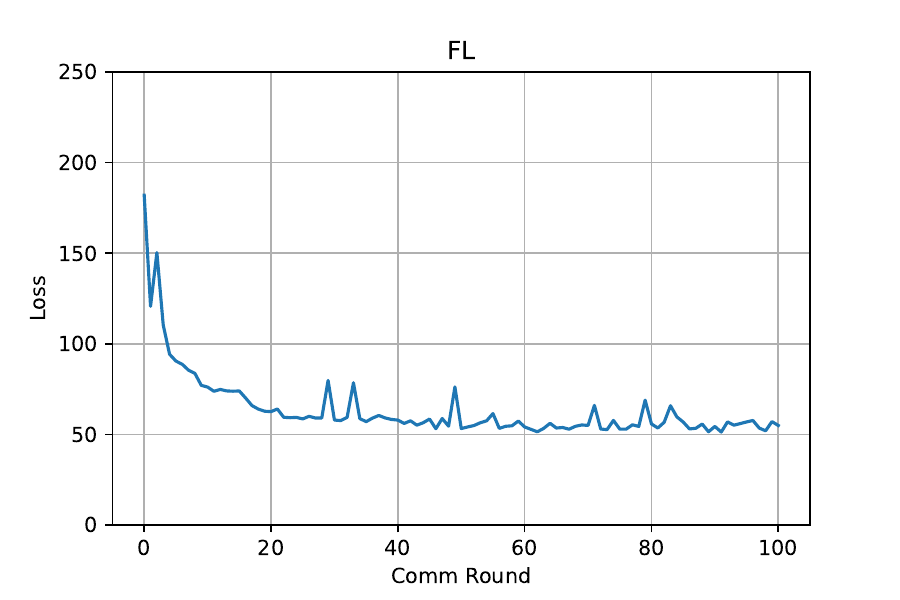}
    \end{subfigure}
    \caption{Loss evolution of CNL (left) and FL (right) systems in the CIFAR10 dataset.}
    \label{fig:FL_CIFAR}
\end{figure*}

CNL slightly outperforms the rest of the systems. This is reasonable as it uses the whole dataset and does not have to deal with the challenges associated to distributed systems. CNL achieves 0.989 and 0.789 accuracy in MNIST and CIFAR10, respectively. Moreover, its loss evolution, shown in Figures~\ref{fig:FL_MNIST} (MNIST) and \ref{fig:FL_CIFAR} (CIFAR10), is two orders of magnitude below the distributed approaches; denoting lower initial convergence challenges.

Regarding the FL systems, MNIST 10-agent scenario achieves an accuracy of 0.985 after 24 communication rounds. Figure \ref{fig:FL_MNIST} shows a fast convergence rate during the evaluation stage of each communication round. On the other hand, in the CIFAR10 dataset, convergence rate is slower, but still relatively good results are achieved: 0.791 accuracy after 101 communication rounds. This is reasonable because CIFAR10 is more complex than MNIST in terms of color channels, size and diversity of the images. The larger, 20-agent, scenarios still feature competitive results which show the good scalability of the FL models. As shown in Table \ref{table_res}, accuracies achieved in the MNIST dataset are 0.986 and the ones achieved in the CIFAR10 dataset are 0.778. Overall, FL results are similar, if slightly lower, to those of CNL. 

\subsection{GLow}


\begin{figure*}[!ht]
    \centering
    \begin{subfigure}
        \centering
        \includegraphics[width = .45\textwidth]{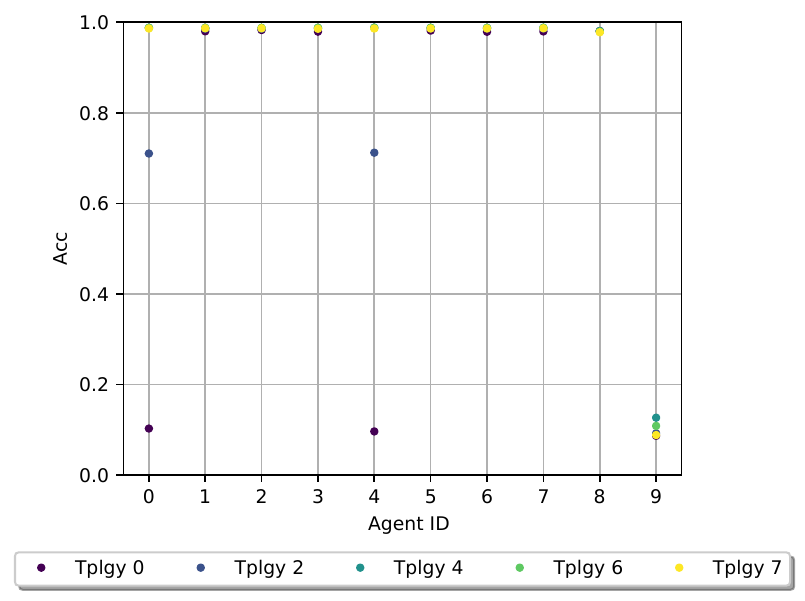}
   \end{subfigure}
    \begin{subfigure}
        \centering
        \includegraphics[width = .45\textwidth]{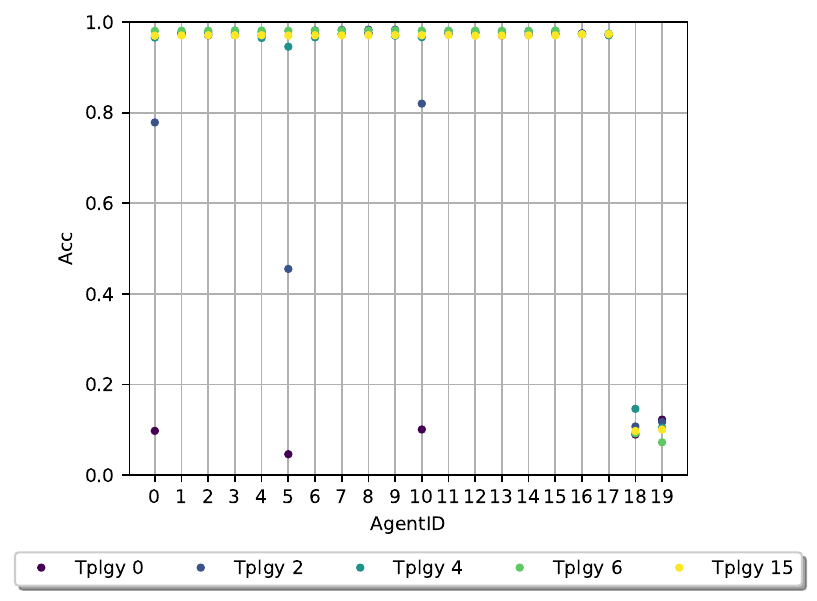}
    \end{subfigure}
    \caption{Per-agent accuracy obtained in GLow simulation, MNIST dataset with 8+2 (left) and 16+4 agents (right).}
    \label{fig:GL_MNIST}
\end{figure*}

Figure \ref{fig:GL_MNIST} shows final accuracies obtained by each agent in the evaluation stage, after running each system for 24 communication rounds in the MNIST dataset -- for both 8+2 and 16+4 scenarios. 
In the 8+2 agent scenario, in topology 0, E agents are not capable of learning any model and act as random guessers. In topology 2, E agents manage to learn information from their neighbors and are able to obtain an accuracy below 0.7 in the final stage of the training -- ED agent 9 is always disconnected from the network acting as a random guesser. In the rest of the topologies (4, 6 and 7), E agents converge similarly compared to R agents in the network. Interestingly, further increasing the degree of inter-connectivity does not improve the results any further. Lastly, D agents achieve an accuracy of 0.99 despite being disconnected from the network. 

The 16+4 scenario has a similar behavior, where D and R agents converge to 0.98 accuracy independently to the degree of connectivity -- despite higher variance is observed. As in the previous case, E agents achieve low accuracies in topologies 0 and 2, but converge well in the rest of the topologies (4-15) -- ED agents 18 and 19 act as random guessers. 

\begin{figure*}[!htb]
\centering
  \includegraphics{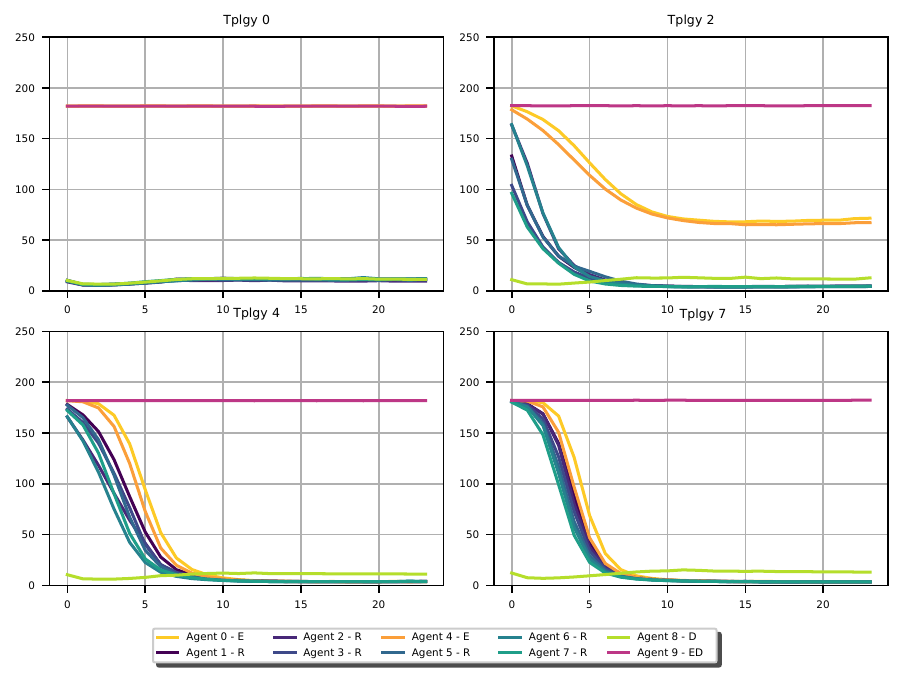}
  \caption{Loss evolution of each 8+2 agents during 24 communication rounds (32 local epochs) for topologies 0, 2, 4 and 7 in MNIST.}
  \label{fig:MNIST_8_32_3}
\end{figure*}

Figure \ref{fig:MNIST_8_32_3} shows the convergence rate of each agent in topologies 0, 2, 4 and 7 in the 8+2 agent scenario (MNIST), showing the loss evolution of each agent per communication round during the evaluation stage: (1) in topology 0 (disconnected graph), E agents do not achieve any convergence, (2) in topology 2, E agents start achieving an observable convergence -- still far from other agents with local data, (3) in topology 4, all connected agents achieve a similar convergence after round 7, (4) in topology 7, the behavior of the agents is similar to topology 4, with a slight boost in convergence -- appreciable in the first 5 rounds of communication, where the loss decrement is more homogeneous. Moreover, agent 8 always performs in SL (with local data) and does not have the initial learning challenge associated to connected agents; which receive parameters from their neighbors. That is translated in a higher initial loss for the agents that are part of the connection graph, but end up converging equally well or better than agent 8 (D). That phenomenon is appreciated between topologies 0 and 2, transitioning from a disconnected graph to a ring. Although initially agent 8 departs from a lower loss, after 7 communication rounds, the loss of agent 8 ends up being slightly over E and R agents -- observable in topologies 2, 4 and 7. 

\begin{figure*}[!ht]
    \centering
    \begin{subfigure}
        \centering
        \includegraphics[width = .45\textwidth]{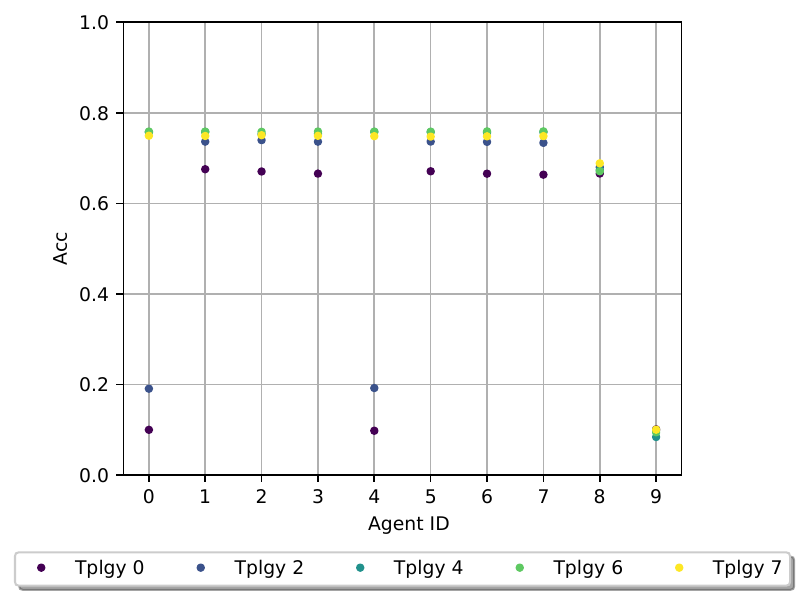}
   \end{subfigure}
    \begin{subfigure}
        \centering
        \includegraphics[width = .45\textwidth]{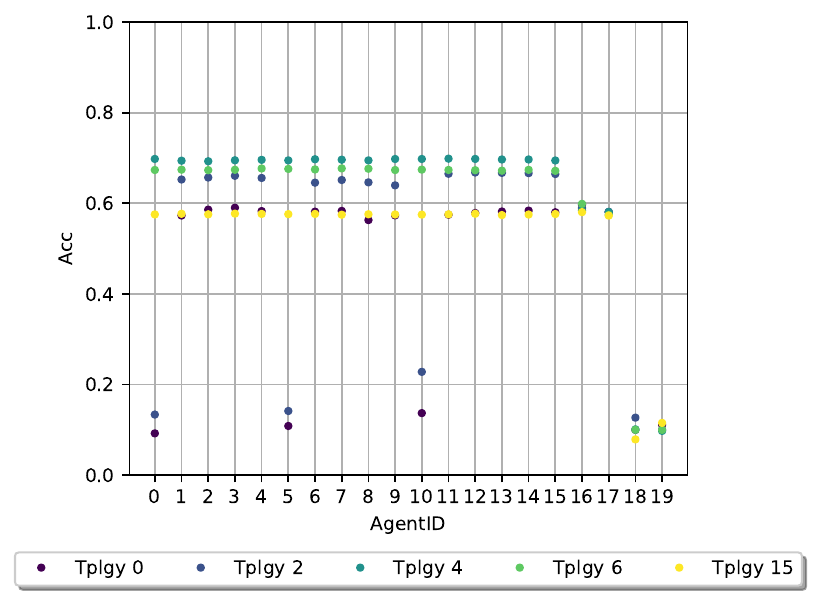}
    \end{subfigure}
    \caption{Per-agent accuracy obtained in GLow simulation, CIFAR10 dataset with 8+2 (left) and 16+4 (right) agents.}
    \label{fig:GL_CIFAR}
\end{figure*}

Next, Figure \ref{fig:GL_CIFAR} shows final accuracies obtained by each agent, in the CIFAR10 dataset (following the same structure of Figure \ref{fig:GL_MNIST}): (1) in the 8+2 agent scenario, a similar behavior to MNIST is observed, E agents do not converge well in topologies 0 and 2 -- in comparison to D and R agents. It is in topologies 4, 6 and 7 where E agents converge to their R neighbors -- agent 9 (ED) is always disconnected from the network acting as a random guesser. Topology 4 is the best performing scenario for all connected agents, obtaining the highest accuracy (average 0.754), independently of whether the agents have local data or not. It is remarkable that being connected (topologies (4-7)) always achieves higher accuracies than the SL scenario of topology 0. In addition, results achieved by special agent 8 (SL) are always below the connected ones, (2) in the 16+4 scenario, similar results are achieved, E agents get a good degree of convergence (accuracy over 0.58) in topologies (4-15) -- whereas ED agents 18 and 19 always perform as random guessers. Moreover, results achieved by connected agents following topologies 4 and 6 show an improvement in comparison to topology 0 (SL), topology 2 (ring) and topologies with a higher degree of connectivity (8-15). Furthermore, results from special D agents 16 and 17 (SL) are always below the connected ones.

\begin{figure*}[!htb]
\centering
  \includegraphics{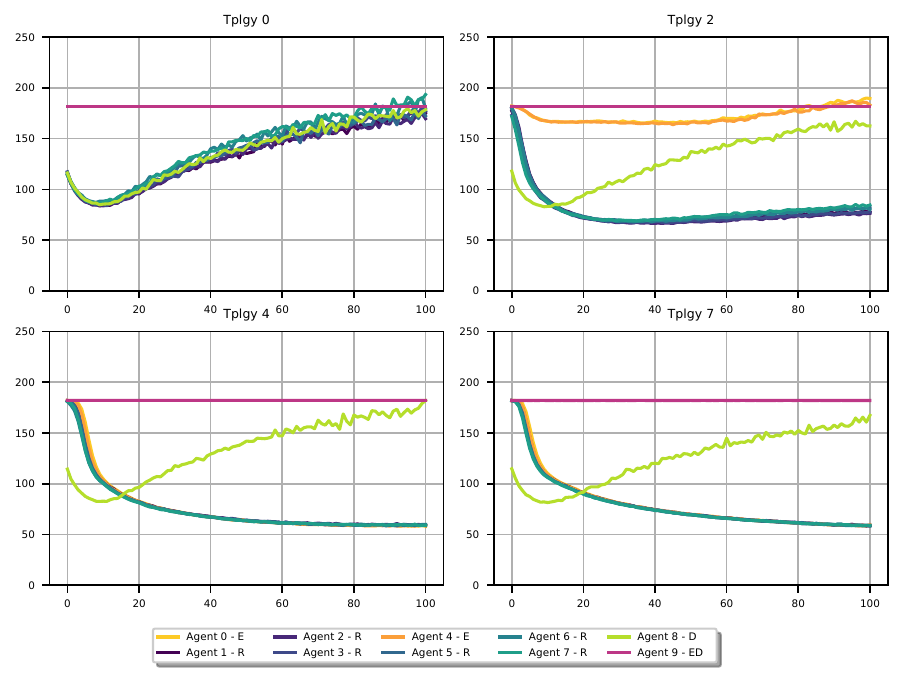}
  \caption{Loss evolution of each 8+2 agents during 101 communication rounds (32 local epochs) for topologies 0, 2, 4 and 7 in CIFAR10.}
  \label{fig:CIFAR_8_32_3}
\end{figure*}

Figure \ref{fig:CIFAR_8_32_3} shows the loss evolution of agents (0-9), in 8+2 scenario, for topologies 0, 2, 4 and 7, during the evaluation stage of 101 communication rounds using CIFAR10 dataset: (1) in topology 0 (disconnected graph), E agents do not achieve any convergence. Other agents containing local data, learn their corresponding SL models achieving their lowest score in the first 15 communication rounds -- loss starts going up afterwards, (2) in topology 2 (ring), E agents start achieving an observable convergence, very far from other agents with local data -- ED agent 9 is always disconnected from the network acting as a random guesser, (3) in topologies 4 and 7, all agents achieve a similar convergence level -- loss of all connected agents decreases in an homogeneous way. Moreover after 40 communication rounds, all connected agents achieve lower losses than when they were disconnected in topology 0. Finally, agent 8 does not achieve a loss level below any interconnected agent -- loss starts going up after communication round 15. 

\subsection{Discussion}

Regarding the accuracies obtained in the MNIST dataset, GLow achieves very similar results to CNL and FL, in spite of GLow being a fully decentralized learning strategy -- as seen in Table \ref{table_res}. On the other hand, focusing on GLow, it is worth mentioning that the expected random guessing behavior of ED agents is fulfilled -- Figure~\ref{fig:GL_MNIST}. This demonstrates the benefits of introducing the feature about special agents for monitoring the behavior of the system. Similarly, agents that were initially disconnected from the network with no local data (E), once they are connected and get information from peers, start improving and, eventually, converging equally well to their neighbors with local instances (R). This happens in both 8+2 and 16+4 scenarios: (1) in the ring topology, E agents start converging and an improvement in their accuracy is observed, but still far from achieving similar results to their neighbors. Aditionally, as seen in Figure \ref{fig:MNIST_8_32_3}, the loss of E agents is far from the rest of R neighbors, (2) with the double ring topology, which provides enough connectivity degree, E agents show a very similar behavior to R agents, (3) this is maintained with more densely connected topologies, but there is not any appreciable benefit in terms of accuracy; the double ring already provides enough connectivity. Indeed, further increasing connectivity has little impact on accuracy but helps speeding up convergence as seen in Figure \ref{fig:MNIST_8_32_3}. The loss of connected agents decreases in less communication rounds in topology 7 than in topology 4. Finally, potential benefits of GLow over SL are observed by looking at agent 8 (D), in the 8+2 scenario: its final loss is slightly higher and its accuracy substantially lower when compared to other interconnected E and R agents in topologies 2, 4 and 7.

Comparing the results achieved in the CIFAR10 dataset by FL, Figure \ref{fig:FL_CIFAR}, and the ones achieved by GLow 8+2, Figure \ref{fig:GL_CIFAR}, a slight difference in performance in favor of FL is observed: 0.791 and 0.754 accuracy respectively -- being the reference connected GLow agents in double ring topology. However, the difference is not very notorious taking into account additional orchestration challenges involved in a fully decentralized system and the handicaps mentioned above. 

Furthermore, due to the complexity of CIFAR10 (e.g., 32$\times$32px images, three color channels and so on), lower accuracy than in the MNIST experiments are observed in all CNL, FL and GLow systems. Even after letting both systems converge for 101 communication rounds, rather than 24. On the other hand, comparing both plots in Figure \ref{fig:GL_CIFAR}, the 8+2 agent scenario achieves better results than the 16+4 one. Connected agents in 8+2 achieve an average accuracy of 0.754 with a variance of 0.00001 for topologies (4-7), whereas connected agents in 16+4 achieve an average accuracy of 0.683 and a variance of 0.0017 for topologies (4-15). We believe this is caused by having less local instances per agent, the existence of disconnected agents and the need of more communication rounds to converge -- as the same dataset is splitted among a larger set of agents and some of them never contribute to the network.

At any rate, the benefits of using GLow against SL are appreciated by comparing the results achieved by interconnected agents (E and R) against the performance achieved by D agents. This is translated into a better performance for all E and R agents in topologies (2-7), over topology 0. Similarly, special agent 8 (D) achieves worse results compared to the rest of interconnected agents in topologies (2-7) -- Figure \ref{fig:GL_CIFAR}, 8+2 scenario. Additionally, results achieved in the 16+4 scenario show the benefits of being interconnected by GLow over SL: accuracies obtained by R agents in topologies 4 and 6 clearly outperform SL (topology 0) and D agents 16, 17.

Regarding the loss evolution in Figure \ref{fig:CIFAR_8_32_3}, the benefits of using GL over SL are observed as well: (1) agents with local data (D and R) in topology 0 never achieve loss levels below 70, whereas much lower values are reached by the rest of GLow topologies, i.e., convergence is boosted by sharing parameters with neighbors, (2) special agent D, always performs worse than other agents in any topology; not benefiting from the parameters of its neighbors. Moreover, in the current model, overfitting is observed in SL agents after communication round 15, when the loss starts going up and an early stopping criteria is not applied.

In respect of the degree of inter-connection, a higher degree of connectivity does not necessarily mean a better performance. As seen in the CIFAR10 experiments, scenario 16+4 in Figure \ref{fig:GL_CIFAR}. Very densely connected topologies (15) achieve slightly lower accuracies than topologies (4-6). This is due to the complexity of CIFAR10, combined with a high number of agents. We believe that increasing the degree of connectivity makes the system converge at slower rates, in scenarios with a large set of agents containing reduced size local datasets. Therefore, a trade-off among the degree of connectivity and the number of agents needs to be taken into consideration when designing GLow simulations. 
\section{Conclusions and Future Work} \label{conclusions}
GLow is a novel, simulation-oriented Gossip Learning strategy built on top of Flower, a state-of-the-art framework. It is completely modular and easy to deploy. We relied on Flower to ensure GLow implementation is community-friendly and open for research purposes. Moreover, the availability of different agent/topology configurations, models and datasets can be done with minimal changes in the code\footnote{Available open source repository: \url{https://github.com/AitorB16/GLow}}. It is worth mentioning that GLow includes a topology generator and result visualization features -- providing a robust tool-set for the study of convergence subject to the degree of inter-connectivity among agents. In addition, the possibility of adding agents operating in SL mode, disconnected from the network (with or without local instances), adds extra flexibility to the evaluation of the system; showing the benefits of performing decentralized parameter averaging among neighbors.


Results obtained in Section~\ref{results} show a good performance of GLow as a Flower built fully decentralized GL strategy. In the 8+2 agent scenario, the three systems achieve very similar accuracies in both MNIST and CIFAR10 datasets -- average GLow accuracies in MNIST: 0.987, and CIFAR10: 0.754. On the other hand, in the 16+4 scenario, GLow obtains similar results to CNL and FL in the MNIST dataset -- 0.971 average accuracy. Nevertheless, slightly lower accuracies are achieved by GLow in the 16+4 CIFAR10 scenario -- 0.683 average accuracy. Arguably, this happens as a consequence of splitting the dataset among a larger set of agents, where some of them are disconnected (operating in SL); not contributing to their neighbors. However, the benefits of using a fully decentralized learning strategy against a centralized one (vanilla FL) which requires an aggregation authority are worth underlining. Specifically, in terms of scalability and fault tolerance, where the FL orchestration server is, potentially, the bottleneck and a single point of failure. Therefore, GLow gains special interest in the current circumstance, where the research on GL is scarce due to design and implementation challenges. The availability of such a tool in the community will have enormous benefits to assist delivering decentralized AI to IoT devices.

Finally, fully decentralized learning entails a series of challenges to be tackled in future GLow versions. On the one hand, dealing with non-IID scenarios adds extra convergence challenges that will need to be addressed by attention mechanisms~\cite{Liu2021} or agent clusterization~\cite{clusterization} techniques. On the other hand, current GLow simulation is designed to sequentially select an agent as head -- performing local training and parameter aggregation with its neighbors. In particular, we have chosen round-robin as our head selection algorithm because it well-balances the selection of network agents in each communication round. In the future, other head selection algorithms such as full-random, or priority based ones will be used and their impact on the learning process analyzed. Similarly, a parallel version of GLow, capable of selecting multiple agents as heads, performing simultaneous training-aggregation tasks is projected. Finally, dealing with IoT devices involves the creation of lightweight models to be trained in limited resource environments. The usage of quantization~\cite{quantization} techniques, is an interesting area of research that combined with fully decentralized learning will open new possibilities.
\section*{Acknowledgments}
\noindent  This research work is supported by the Basque Government through projects KK-2023/00012, KK-2023/00090, KK-2024/00030, KK-2024/00068 and IT1504-22. 
It is also supported by grant CNS2023-144315 funded by the MICIU/AEI/10.13039/501100011033 and by ``European Union NextGenerationEU/PRTR''.
It is also supported by grant PID2023-152390NB-I00 funded by the MICIU/AEI/10. 13039/501100011033 and by ``FEDER funds''.
Dr. Javier Navaridas holds a Ram\'on y Cajal fellowship funded by the Spanish Ministry of Science, Innovation and Universities (RYC2018-024829-I), funded by MCIN/AEI/ 10.13039/ 501100011033 and, as appropriate, by ``ESF Investing in your future'' or by ``European Union NextGenerationEU/ PRTR''.
Aitor Belenguer is supported by a PhD fellowship from the Basque Government.

\bibliographystyle{IEEEtran}
\bibliography{main}

\begin{thebibliography}{10}
\providecommand{\url}[1]{#1}
\csname url@samestyle\endcsname
\providecommand{\newblock}{\relax}
\providecommand{\bibinfo}[2]{#2}
\providecommand{\BIBentrySTDinterwordspacing}{\spaceskip=0pt\relax}
\providecommand{\BIBentryALTinterwordstretchfactor}{4}
\providecommand{\BIBentryALTinterwordspacing}{\spaceskip=\fontdimen2\font plus
\BIBentryALTinterwordstretchfactor\fontdimen3\font minus \fontdimen4\font\relax}
\providecommand{\BIBforeignlanguage}[2]{{%
\expandafter\ifx\csname l@#1\endcsname\relax
\typeout{** WARNING: IEEEtran.bst: No hyphenation pattern has been}%
\typeout{** loaded for the language `#1'. Using the pattern for}%
\typeout{** the default language instead.}%
\else
\language=\csname l@#1\endcsname
\fi
#2}}
\providecommand{\BIBdecl}{\relax}
\BIBdecl

\bibitem{BrendanMcMahan2017}
H.~{Brendan McMahan}, E.~Moore, D.~Ramage, S.~Hampson, and B.~{Ag{\"{u}}era y Arcas}, ``{Communication-efficient learning of deep networks from decentralized data},'' \emph{Proceedings of the 20th International Conference on Artificial Intelligence and Statistics, AISTATS 2017}, vol.~54, 2017.

\bibitem{decentr_gl}
I.~Hegedűs, G.~Danner, and M.~Jelasity, ``Decentralized learning works: An empirical comparison of gossip learning and federated learning,'' \emph{Journal of Parallel and Distributed Computing}, vol. 148, pp. 109--124, 2 2021.

\bibitem{flower}
\BIBentryALTinterwordspacing
D.~J. Beutel, T.~Topal, A.~Mathur, X.~Qiu, T.~Parcollet, and N.~D. Lane, ``Flower: {A} friendly federated learning research framework,'' \emph{CoRR}, vol. abs/2007.14390, 2020. [Online]. Available: \url{https://arxiv.org/abs/2007.14390}
\BIBentrySTDinterwordspacing

\bibitem{borja}
B.~Molina-Coronado, U.~Mori, A.~Mendiburu, and J.~Miguel-Alonso, ``Survey of network intrusion detection methods from the perspective of the knowledge discovery in databases process,'' \emph{IEEE Transactions on Network and Service Management}, vol.~17, no.~4, pp. 2451--2479, 2020.

\bibitem{surv2}
\BIBentryALTinterwordspacing
Q.~Li, Z.~Wen, Z.~Wu, S.~Hu, N.~Wang, Y.~Li, X.~Liu, and B.~He, ``{A Survey on Federated Learning Systems: Vision, Hype and Reality for Data Privacy and Protection},'' pp. 1--44, 2019. [Online]. Available: \url{http://arxiv.org/abs/1907.09693}
\BIBentrySTDinterwordspacing

\bibitem{S48}
C.~Dwork, F.~McSherry, K.~Nissim, and A.~Smith, ``Calibrating noise to sensitivity in private data analysis,'' vol. Vol. 3876, 01 2006, pp. 265--284.

\bibitem{Liu2021}
Y.~Liu, S.~Garg, J.~Nie, Y.~Zhang, Z.~Xiong, J.~Kang, and M.~S. Hossain, ``{Deep Anomaly Detection for Time-Series Data in Industrial IoT: A Communication-Efficient On-Device Federated Learning Approach},'' \emph{IEEE Internet of Things Journal}, vol.~8, no.~8, pp. 6348--6358, 2021.

\bibitem{ormandi}
R.~Ormándi, I.~Hegedus, and M.~Jelasity, ``Gossip learning with linear models on fully distributed data,'' in \emph{Concurrency and Computation: Practice and Experience}, vol.~25.\hskip 1em plus 0.5em minus 0.4em\relax John Wiley and Sons Ltd, 2 2013, pp. 556--571.

\bibitem{Yuan2023}
\BIBentryALTinterwordspacing
L.~Yuan, Z.~Wang, L.~Sun, P.~S. Yu, and C.~G. Brinton, ``Decentralized federated learning: A survey and perspective,'' 6 2023. [Online]. Available: \url{http://arxiv.org/abs/2306.01603}
\BIBentrySTDinterwordspacing

\bibitem{mbeltran}
E.~T. Martínez~Beltrán, M.~Q. Pérez, P.~M.~S. Sánchez, S.~L. Bernal, G.~Bovet, M.~G. Pérez, G.~M. Pérez, and A.~H. Celdrán, ``Decentralized federated learning: Fundamentals, state of the art, frameworks, trends, and challenges,'' \emph{IEEE Communications Surveys \& Tutorials}, vol.~25, no.~4, pp. 2983--3013, 2023.

\bibitem{braintorrent}
\BIBentryALTinterwordspacing
A.~G. Roy, S.~Siddiqui, S.~Pölsterl, N.~Navab, and C.~Wachinger, ``Braintorrent: A peer-to-peer environment for decentralized federated learning,'' 5 2019. [Online]. Available: \url{http://arxiv.org/abs/1905.06731}
\BIBentrySTDinterwordspacing

\bibitem{fedstellar}
E.~T.~M. Beltrán, Ángel Luis Perales~Gómez, C.~Feng, P.~M.~S. Sánchez, S.~L. Bernal, G.~Bovet, M.~G. Pérez, G.~M. Pérez, and A.~H. Celdrán, ``Fedstellar: A platform for decentralized federated learning,'' \emph{Expert Systems with Applications}, vol. 242, 5 2024.

\bibitem{Chen2024}
L.~Chen, W.~Liu, Y.~Chen, and W.~Wang, ``Communication-efficient design for quantized decentralized federated learning,'' \emph{IEEE Transactions on Signal Processing}, vol.~72, pp. 1175--1188, 2024.

\bibitem{gossipfl}
Z.~Tang, S.~Shi, B.~Li, and X.~Chu, ``Gossipfl: A decentralized federated learning framework with sparsified and adaptive communication,'' \emph{IEEE Transactions on Parallel and Distributed Systems}, vol.~34, pp. 909--922, 3 2023.

\bibitem{Kanamori2023}
Y.~Kanamori, Y.~Yamasaki, S.~Hosoai, H.~Nakamura, and H.~Takase, ``An asynchronous federated learning focusing on updated models for decentralized systems with a practical framework,'' in \emph{Proceedings - International Computer Software and Applications Conference}, vol. 2023-June.\hskip 1em plus 0.5em minus 0.4em\relax IEEE Computer Society, 2023, pp. 1147--1154.

\bibitem{MNIST}
\BIBentryALTinterwordspacing
Y.~LeCun and C.~Cortes, ``{MNIST} handwritten digit database,'' 2010. [Online]. Available: \url{http://yann.lecun.com/exdb/mnist/}
\BIBentrySTDinterwordspacing

\bibitem{cifar}
\BIBentryALTinterwordspacing
A.~Krizhevsky, V.~Nair, and G.~Hinton, ``Cifar-10 (canadian institute for advanced research).'' [Online]. Available: \url{http://www.cs.toronto.edu/~kriz/cifar.html}
\BIBentrySTDinterwordspacing

\bibitem{le_net}
Y.~Lecun, L.~Bottou, Y.~Bengio, and P.~Haffner, ``Gradient-based learning applied to document recognition,'' \emph{Proceedings of the IEEE}, vol.~86, no.~11, pp. 2278--2324, 1998.

\bibitem{clusterization}
W.~Lai, Z.~Xu, and Q.~Yan, ``Clustered federated learning based on client’s prototypes,'' in \emph{2024 27th International Conference on Computer Supported Cooperative Work in Design (CSCWD)}, 2024, pp. 909--914.

\bibitem{quantization}
U.~Kulkarni, A.~S. Hosamani, A.~S. Masur, S.~Hegde, G.~R. Vernekar, and K.~Siri~Chandana, ``A survey on quantization methods for optimization of deep neural networks,'' in \emph{2022 International Conference on Automation, Computing and Renewable Systems (ICACRS)}, 2022, pp. 827--834.

\end{thebibliography}



\end{document}